\documentclass[lettersize,journal]{IEEEtran}
\usepackage{amsmath,amsfonts}
\usepackage{algorithmic}
\usepackage{algorithm}
\usepackage{array}
\usepackage[caption=false,font=normalsize,labelfont=sf,textfont=sf]{subfig}
\usepackage{textcomp}
\usepackage{stfloats}
\usepackage{url}
\usepackage{verbatim}
\usepackage{graphicx}
\usepackage{xurl}
\usepackage{cite}
\usepackage[utf8]{inputenc} 
\usepackage[T1]{fontenc}    
\usepackage{hyperref}       
\usepackage{url}            
\usepackage{booktabs}       
\usepackage{amsfonts}       
\usepackage{nicefrac}
\usepackage{graphicx}
\usepackage{microtype}      
\usepackage{xcolor} 
\usepackage[numbers]{natbib}

\begin{document}

\title{Signal from Space: Detecting Schools and Towers to Bridge the Digital Divide}


\author{Zakarya Elmimouni \IEEEauthorrefmark{1},
Sandor Farkas \IEEEauthorrefmark{2}, 
Fares Fourati \IEEEauthorrefmark{1},
Vladimir Daigele \IEEEauthorrefmark{2},
Walid Mathlouthi \IEEEauthorrefmark{2}, 
Mohamed-Slim Alouini \IEEEauthorrefmark{1}

\thanks{\IEEEauthorrefmark{1}King Abdullah University of Science and Technology (KAUST), Thuwal, Saudi Arabia.}
\thanks{\IEEEauthorrefmark{2}International Telecommunication Union (ITU), Geneva, Switzerland.}
\thanks{Emails: zakarya.elmimouni@gmail.com, \{fares.fourati, slim.alouini\}@kaust.
edu.sa \{sandor.farkas, vladimir.daigele, walid.mathlouthi\}@itu.int \textbf{Corresponding author: Zakarya Elmimouni} }


}



\maketitle

\begin{abstract}

Reliable internet access is essential for modern education, yet millions of school‑aged children especially in developing regions remain offline due to unconnected schools. The Giga Initiative aims to connect every school to the internet, but doing so at scale requires efficient methods to map schools and assess surrounding connectivity infrastructure without relying on sparse or noisy third‑party datasets. In this work, we propose a scalable, vision‑only framework that uses high‑resolution satellite imagery and transfer learning to address both tasks simultaneously. By adapting pre‑trained object detection models to new geographical regions with minimal labeled data, we detect schools and cellular towers directly from space. We then analyze the spatial relationship between detected schools and nearby towers as a proxy for connectivity availability . This purely imagery‑driven pipeline enables large‑scale infrastructure mapping, reduces dependency on auxiliary data, and supports data‑driven prioritization of connectivity investments in underserved areas. Our approach is demonstrated on real satellite imagery from Lesotho, showing strong performance across this region.

\end{abstract}


\section{Introduction}

The global learning crisis faces an urgent inflection point: nearly 2.2 billion people remain offline \cite{itu2025}, with children in low-income countries disproportionately excluded from digital opportunity \cite{UNICEF2020}. In Sub-Saharan Africa, the situation is particularly alarming: a study by UNICEF and the International Telecommunication Union (ITU) shows that 95\% of school-age children in West and Central Africa (194 million) and 88\% in East and Southern Africa (191 million) lack internet access at home \cite{UNICEFITU2020}. In Latin America and the Caribbean, 49\% of children aged 3 to 17 remain unconnected. These stark regional disparities underscore the urgency of bridging the digital divide to guarantee inclusive and equitable education for all \cite{UNICEFITU2020}.

In response, UNICEF and the ITU launched the Giga Initiative, aiming to connect every school to the internet by 2030 \cite{giga2025}. Giga's approach rests on three interdependent pillars: first, mapping the precise locations of schools; second, modeling the surrounding infrastructure including cellular towers and fiber routes that can support connectivity; and third, using these maps to plan and finance the deployment of internet services to unconnected schools. Giga's early tool, Project Connect, evolved into Giga Maps, which today shows more than 2.2 million schools across 140 countries \cite{giga2025}.

To support such large‑scale connectivity planning, the International Telecommunication Union (ITU), in partnership with Ericsson, has developed the Connectivity Planning Platform (CPP)  a geographic information system‑based tool designed to help decision‑makers visualise, analyse, and address infrastructure gaps using multi‑source geospatial data, population distribution, and terrain elevation models \cite{itu_ericsson_cpp_2025}. The CPP enables evidence‑based assessments of potential point‑to‑point (P2P) links, making it particularly well‑suited for evaluating whether a detected school can realistically be connected to a nearby cellular tower. In this work, we leverage the CPP’s P2P radio visibility analysis model as the core of our connectivity inference pipeline, moving beyond simple geodesic distance to a terrain‑aware, engineering‑grade assessment.

Existing methods for predicting school connectivity often fall short of real world applicability due to their reliance on the availability, density, and accuracy of third‑party datasets. For example, Doerksen et al.\cite{doerksen2024} depend on extensive secondary features including medium‑resolution satellite indicators (nighttime lights, land cover), proximity to power lines, and crowdsourced network performance metrics such as Ookla speed tests, which are frequently sparse or noisy in underserved regions. This dependency fundamentally limits the scalability and robustness of their approach where such data are missing or unreliable.

In this work, we overcome these limitations by proposing a two‑level methodology to assess school connectivity directly from space, using only satellite imagery. Our approach builds on a scalable school detection pipeline introduced in \cite{our_work} and a complementary cell tower detection model adapted for Mozambique. First, we detect schools from high‑resolution satellite imagery by fine‑tuning a previously trained model on our small, manually curated dataset. Second, we apply an oriented bounding box detector (OBB) to locate cell towers in the same region. We then analyze the spatial relationship between each detected school and its nearest tower, using the ITU CPP P2P radio visibility analysis model as the core of our connectivity assessment 
\cite{ituP2P}. This framework directly models a potential point-to-point microwave link between a school and a tower by processing the geodesic (true earth surface) distance and the intervening terrain elevation to determine true line-of-sight. This terrain-aware P2P model provides a realistic, engineering-grade assessment of whether a usable signal can be established, moving beyond simple planar distance proxies. This two-stage, purely vision-driven pipeline enables large-scale, data-independent identification of unconnected schools and directly supports infrastructure deployment prioritization in underserved areas such as Lesotho.

\section{Related Work}

Object detection is a core task in computer vision, involving both the identification and localization of specific object categories within complex visual scenes \cite{PAGIRE2025110075}. It has been widely applied across domains such as agriculture \cite{badgujar2024agricultural}, medical applications \cite{ragab2024comprehensive}, and remote sensing \cite{zhu2024scnet, xiao2025yolors, Sirko2021Continental}. Recent advances in detection architectures, including Grounding Dino \cite{liu2024grounding}, DETR \cite{detr}, several YOLO-based models \cite{yolov10, yolo11_ultralytics, yolo12_ultralytics} and RCNN based models \cite{rcnn,fastrcnn,fasterrcnn} have significantly improved both accuracy and inference speed, enabling large-scale deployment in real-world settings.

School mapping from aerial and satellite imagery has traditionally been approached as a tile-level classification task rather than explicit object detection \cite{Yazdani2018,yi2019,maduako2022,doerksen2024}. While CNN-based classifiers can effectively identify tiles likely to contain schools, they lack the precise bounding-box localization required for infrastructure planning. To mitigate this,\cite{tingzon2025} proposed approximating school locations using Class Activation Maps (CAMs) derived from classification models. However, CAMs are inherently limited by model interpretability and reliability, as highlighted regions do not consistently align with actual school buildings\cite{Amin2023}. More recently,\cite{our_work} reformulated the task as a weakly supervised object detection problem, demonstrating that high-quality detection can be achieved with only 50--100 labeled images by leveraging GPS coordinates and segmentation priors. Our school detection pipeline builds directly on this label-efficient paradigm.

Complementary to school mapping, automated detection of telecommunication infrastructure from satellite imagery has emerged as a critical enabler for connectivity planning in underserved regions. Recent work by \cite{Krell2023} demonstrates a partially automated workflow for cell tower detection across 26 African countries, leveraging deep neural networks trained on high-resolution  imagery and OpenStreetMap-derived annotations to achieve good detection performance. This directly addresses the data scarcity challenge faced by initiatives like Giga, where manual infrastructure mapping is infeasible at scale. 


Beyond locating facilities, assessing their digital connectivity is a critical prerequisite for bridging the digital divide. Determining this status is challenging, as official telecommunication records are often non-existent or inaccurate in developing regions. Recent efforts, notably by \cite{doerksen2024}, address school connectivity prediction by framing it as a binary classification problem based on multimodal tabular data. Their methodology profiles each school using extensive secondary features incorporating medium-resolution satellite indicators (nighttime lights, land cover), proximity to power lines, and crowdsourced network performance metrics such as Ookla speed tests, to train traditional machine learning classifiers like Random Forests and Support Vector Machines. While analytically comprehensive, this methodology is heavily dependent on the availability, density, and accuracy of third-party datasets, which are frequently sparse or noisy in underserved areas. To circumvent this dependency, our framework evaluates connectivity infrastructure directly from high-resolution imagery. By deploying the complementary object detection model to locate cell towers, we analyze the spatial distance between each school and the nearest tower. This establishes a purely vision-driven proxy for connectivity availability, enabling a more autonomous and scalable assessment independent of auxiliary tabular data.    

\begin{figure*}[t]
    \centering
    \includegraphics[width=\textwidth]{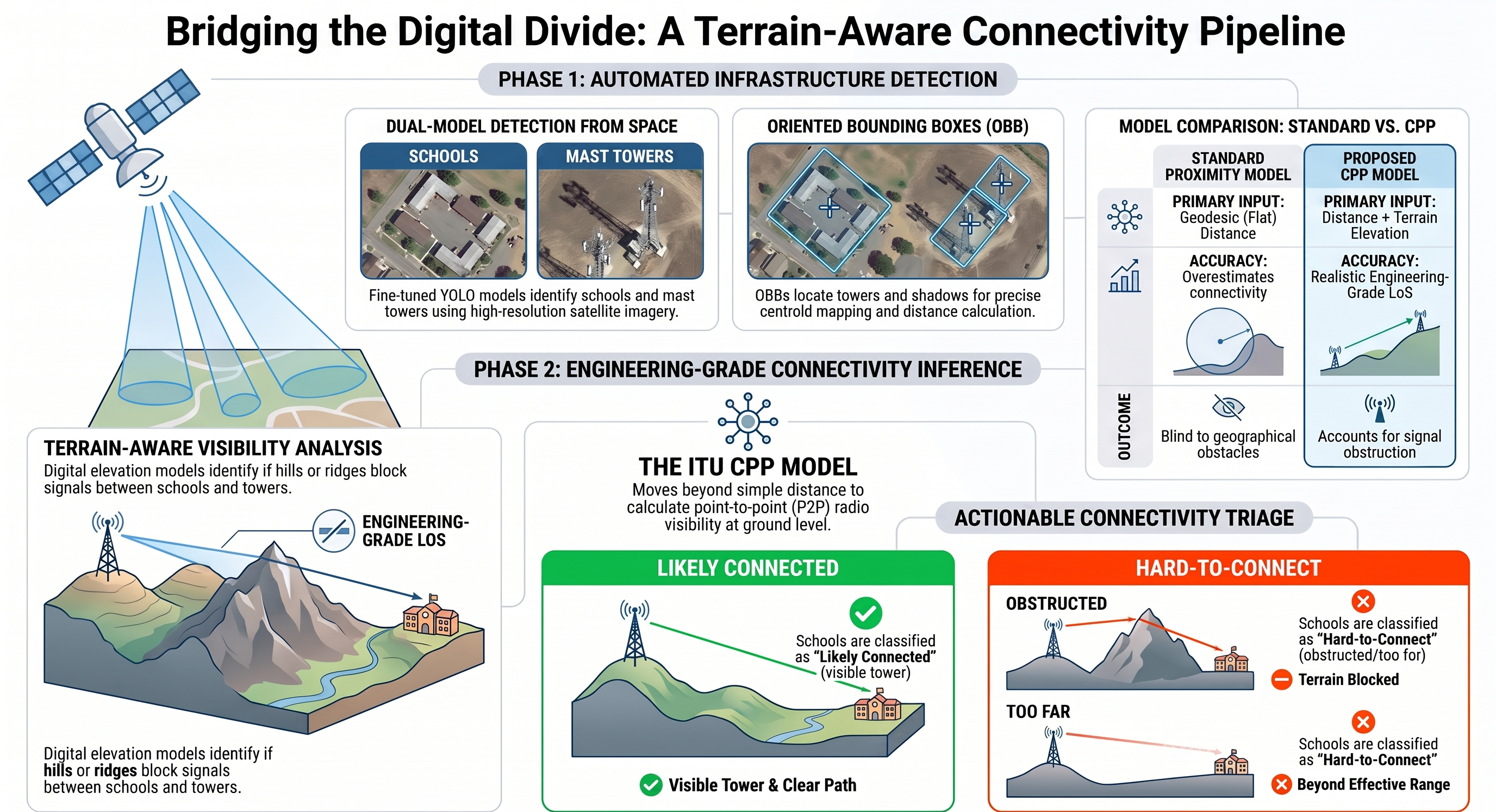}
    \caption{\small \textbf{Inference Pipeline for School Detection and Connectivity Assessment.} The process consists of three main stages: (1) \textbf{Detection of schools} over a large area using sliding search windows on satellite imagery \cite{our_work};(2) \textbf{Tower search, line-of-sight visibility analysis}, and ground distance calculation within a 1km connectivity zone to identify the nearest visible cell tower; and (3) \textbf{generation of a priority map} and connectivity list to identify schools for investment based on their connectivity status.}
    \label{fig:school_connectivity_pipeline}
\end{figure*}

\section{Datasets Construction}

\subsection{School and non School Dataset}
We construct a manually annotated dataset for school detection in Lesotho. Positive samples (schools) are sourced from geolocated coordinates provided by the GIGA initiative~\cite{giga_data}. To capture realistic false positives, negative samples are collected from OpenStreetMap (OSM) features corresponding to visually similar non-school infrastructure (hospitals, commercial buildings, parking lots) exclusively in built-up areas. After removing duplicates and enforcing a minimum $150$~m separation to reduce spatial overlap, we download $512 \times 512$ pixel tiles at $\sim 0.6$~m resolution. From this collection, we manually curate a high-quality ``golden'' subset containing  school and non-school images. Precise bounding boxes are annotated for all visible school buildings. This manually labeled dataset is partitioned into 200 training, 93 validation, and 129 test images.

\subsection{Cell Tower Dataset}
To assess connectivity infrastructure, we constructed a dedicated dataset for cell towers in Lesotho. The dataset comprises 179 high-resolution satellite tiles manually annotated with oriented bounding boxes. We focused on two classes: \textit{Type-1 towers (mast towers)} ($n=188$) and their corresponding \textit{shadows}  \color{black}. Shadows are explicitly annotated as they provide robust spatial cues for vertically slender structures that are often visually ambiguous in top-down imagery. The dataset was randomly partitioned into training (70\%), validation (10\%), and test (20\%) splits. Background and negative samples are naturally represented across the tile collection, and hard negatives are implicitly handled during training via standard data augmentation and the sliding-window inference protocol.

\section{Methodology}
Our overall pipeline consists of three main stages: (1) school detection, (2) cell tower detection, and (3) connectivity inference based on spatial proximity. Figure~\ref{fig:school_connectivity_pipeline} provides an overview of the proposed approach. In the following subsections, we detail the model selection and fine-tuning strategy for school detection, briefly reference the parallel work on tower detection, and then describe the combined inference procedure.

\subsection{School Detection Model Selection and Fine-Tuning}

We benchmark several approaches using YOLO26n on our manually annotated golden dataset. The different approaches include:

\begin{itemize}
    \item  YOLO26n directly trained on the golden dataset.
    
    \item Zero-shot inference using the best YOLO model trained over the USA \cite{elmimouni2026code}.
    
    \item Finetune The best YOLO26n model pre-trained on the USA golden dataset \cite{elmimouni2026code}.
    
    
\end{itemize}

We use transfer learning to adapt the models to the specific visual characteristics of schools in Lesotho. Fine-tuning hyperparameters are optimized using ECP hyper-parametrization on the validation set \cite{fourati2025every, fourati26ecpv2}. The primary evaluation metric is the mean Average Precision (mAP50) and the (mAP@0.5:0.95), along with F1 score and precision and  recall.

After benchmarking, we select the model that achieves the highest performance on the test set as our final school detector. This model is then frozen and used in the inference pipeline.

\subsection{Cell Tower Detection Model}
Complementing the school detector, we deploy an Oriented Bounding Box variant of YOLO26 \cite{ultralytics_obb_docs} to localize cellular infrastructure. OBBs are preferred over axis-aligned boxes as they minimize background interference and yield tighter centroid localization for distance computation. We initialize the detector with weights locally pre-trained on a regional dataset from Mozambique, which provides robust feature representations for narrow vertical structures and elongated shadow patterns.

\subsection{Integrated Inference Pipeline for Connectivity Assessment}

Once both detectors are trained, we combine them into a two‑stage inference pipeline that operates at the scale of a selected area in Mafeteng province in Lesotho as shown in Figure \ref{fig:research_area}. Figure~\ref{fig:school_connectivity_pipeline} provides an overview of the whole pipeline. it proceeds as follows:

\begin{figure}[t]
    \centering
    \includegraphics[width=\columnwidth]{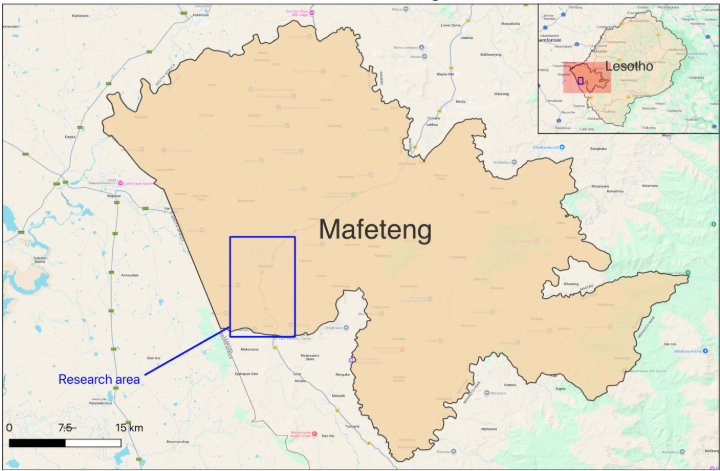}
    \caption{\small Spatial Extent of the Research Area in Mafeteng District, Lesotho}
    \label{fig:research_area}
\end{figure}

\paragraph{Stage 1: School detection over the entire region}
We divide the region of interest into overlapping tiles of fixed size ( \(512 \times 512\) pixels at 0.6~m resolution). A sliding window stride of \(256\) pixels is used to ensure coverage of schools that may lie on tile boundaries. Each tile is passed through our school detector. For every detected school, we record its centroid coordinates \((x_s, y_s)\) and the detection confidence score. Duplicate detections of the same school (across overlapping tiles) are removed . The output of this stage is a unique set of school locations across the entire region.

\paragraph {Stage 2: Cell tower detection and Connectivity assessment}
For each detected school, we detect cell towers within a $1\,km$ radius around the school. Rather than relying on simple geodesic distance between school and tower coordinates, we use the ITU CPP P2P radio visibility analysis model, which incorporates a digital terrain elevation model to determine whether an unobstructed line-of-sight actually exists between each school-tower pair at ground level. This distinction matters in practice: two schools located at an identical geodesic distance from a tower can have markedly different connectivity prospects if intervening terrain (hills, ridges...) blocks the signal path for one but not the other. A distance-only proxy would classify both schools identically, despite one being physically unreachable via a direct radio link. By first filtering candidate towers on visibility and then computing ground distance to the nearest \textit{visible} tower, CPP provides a more realistic, terrain-aware estimate of which schools can plausibly be served by existing infrastructure, rather than relying on proximity alone.

Concretely, the detected school positions and cell tower positions are used as inputs to the CPP P2P radio visibility analysis model, which returns a visibility flag and the ground distance for each school-tower pair. We consider a school \textit{to have high connectivity potential} when at least one tower with confirmed line-of-sight is found within the 1km search radius. When no such visible tower is found, the school is labeled as \textit{hard-to-connect}, indicating that, based on  visible infrastructure, it is unlikely to be served without further investment.

The search radius (1~km) and the distance thresholds can be adjusted based on local network operator data and terrain complexity.




\begin{figure*}[t]
    \centering
    \includegraphics[width=0.8\textwidth]{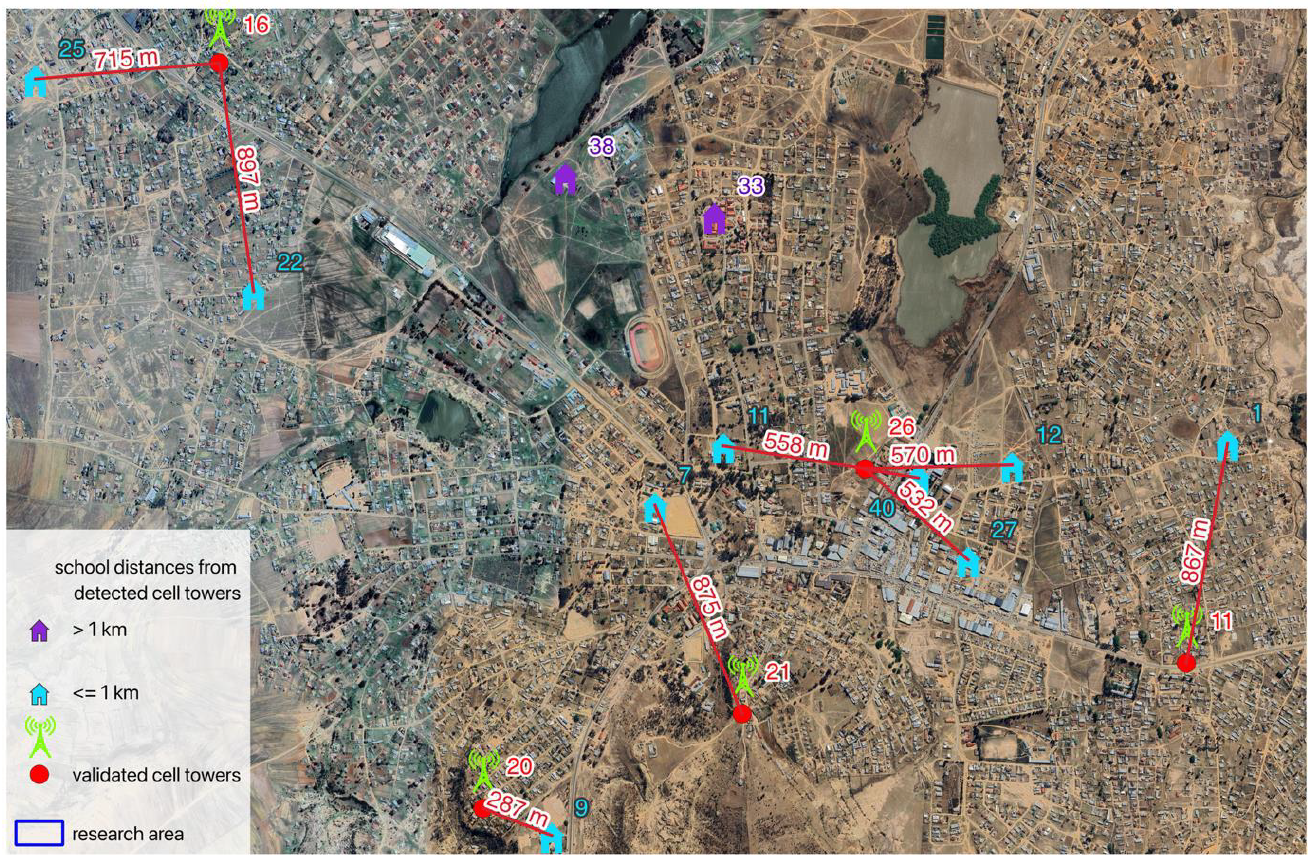}
    \caption{\small  Geographic Distribution of Schools and Cell Towers with Confirmed Line-of-Sight, Highlighting Likely Connectable and Hard-to-Connect Schools.}
    \label{fig:school_proximity_map}
\end{figure*}

\section{Results}
All training and fine-tuning experiments were performed using  NVIDIA GTX 1080Ti GPUs. Fine-tuning on the golden datasets was complemented by automatic hyperparameter optimization using the ECP framework \cite{fourati2025every, fourati26ecpv2}.

For school detection, we compare the three training strategies described in the methodology section. The corresponding results are presented in Table \ref{table:lesotho_results}.




\begin{table}[t]
\centering
\begin{tabular}{lccccc}
\toprule
\textbf{Metric} & \textbf{Model1} & \textbf{Model2} & \textbf{Model3} \\
\midrule
mAP@50        & 0.399 & 0.899 & \textbf{0.913} \\
Precision@50  & 0.302 & 0.880 & 0.840\\
Recall@50     & 0.470 & 0.884& \textbf{0.951}\\
F1@50         & 0.368 & 0.882 & \textbf{0.892}\\
mAP@50:95     & 0.203 & 0.536 & \textbf{0.536}\\
\bottomrule
\end{tabular}

\caption{\small Evaluation on Lesotho's manually labeled test set. Model1 reports the zero-shot predictions of the best YOLO26n model trained on the USA in \cite{our_work}. Model2 corresponds to YOLO26n trained directly on the golden training set. Model3 is the best USA-trained YOLO26n model \cite{our_work} fine-tuned on Lesotho's golden data.}
\label{table:lesotho_results}
\end{table}

Based on these results, we select \textbf{Model 3} (the USA pre-trained detector fine-tuned on Lesotho's golden dataset) for large-scale deployment. Its superior mAP@50 (0.913), F1 score (0.892) and exceptionally high recall (0.951) directly align with our operational objective: in infrastructure mapping, minimizing false negatives is critical, as missing an existing school compromises connectivity planning.

We deployed this detector across a selected area in Mafeteng province in Lesotho with a confidence threshold of 0.6, using the sliding-window protocol described in the previous section. The  automated inference yields $20$ candidate school detections. 

for the cell tower detection, the Mozambique model is fine-tuned on the Lesotho dataset using transfer learning. Training is conducted for 100 epochs with a batch size of 16, input resolution $1024\times1024$. Given the operational priority of minimizing missed infrastructure, the detection pipeline is optimized for high recall. Inference employs a 0.6 confidence threshold. On the test set, the fine-tuned model achieves a mAP@50 of 0.971, Precision@50 of 0.989, Recall@50 of 0.974, and mAP@50:95 of 0.672, demonstrating strong generalization to rural Lesotho topography.

For each school, the OBB-based cell tower detector is applied within a
1\,km radius to assess the surrounding connectivity infrastructure. Schools are classified as \textit{having high connectivity potential} or as \textit{hard-to-connect} based on the output of the CPP P2P radio visibility analysis model, as described in the previous section. As reported in Figure \ref{fig:school_proximity_map} the pipeline identifies $9$  schools with high connectivity potential and $11$ hard-to-connect schools. These results demonstrate the framework's ability to automatically identify connectivity gaps, prioritize underserved schools, and generate actionable maps to support infrastructure planning and investment decisions.


\section{Discussion}
The proposed pipeline is not intended to certify with certainty whether a school is connected to the internet, but rather to provide a vision-derived connectivity potential indicator that gives Giga decision-makers a fast, low-cost, fully automated first-pass screening of areas where ground-truth telecommunication records are sparse or unavailable. 

The "likely-connected" and "hard-to-connect" labels should be read as a triage signal to help prioritize where limited field-verification and investment resources are directed first, complementing rather than replacing human and operator-level validation before any final decision.
A direct consequence of this detection-based framing is that a missed tower will cause any school relying on it to be incorrectly labeled hard-to-connect, with a potential downstream impact on prioritization; we partly mitigate this by tuning both detectors for high recall and by keeping all intermediate outputs auditable against operator registries where available. The pipeline was also demonstrated on a single sub-region with a comparatively small tower dataset, sufficient to validate feasibility but  limited to draw robust conclusions on national-scale gaps or generalization to more diverse terrain, scaling the evaluation across Lesotho and other Giga partner countries is a natural next step before operational deployment.

\section{Conclusion}
In this work, we presented a scalable, fine‑tuning‑driven pipeline to assess school connectivity by jointly detecting schools and cellular infrastructure from high‑resolution satellite imagery. Rather than relying on costly manual annotation or complex weakly supervised labeling, we demonstrate that existing regional detectors (USA‑trained for schools, Mozambique‑trained for towers) can be efficiently adapted to new geographies through fine‑tuning and automated hyperparameter optimization. By explicitly prioritizing high recall, our models minimize false negatives, ensuring that existing infrastructure is reliably captured during large‑scale inference. When deployed across Lesotho’s Mafeteng province, the two‑stage pipeline successfully triaged detected schools into actionable connectivity categories based on point‑to‑point line‑of‑sight analysis using the ITU CPP P2P radio visibility analysis model, demonstrating its practical utility for infrastructure planning.

Our approach addresses a critical bottleneck in global connectivity initiatives by providing a reproducible transfer‑learning workflow that scales across underserved regions without requiring a training from scratch. By adopting the ITU CPP model, we move beyond simple planar distance proxies and incorporate geodesic distance and digital elevation,  to provide an engineering‑grade connectivity assessment. This work offers a foundational step toward data‑driven, equitable internet access in alignment with the Giga Initiative’s 2030 objectives.

\bibliographystyle{unsrtnat}
\bibliography{aaai2026}

\newpage

 




\vfill

\end{document}